\documentclass[10pt, a4paper]{article}
\usepackage{etoolbox}
\usepackage{lrec}
\usepackage{lipsum} 
\usepackage{times}
\usepackage{latexsym}
\usepackage{url}
\usepackage{comment}
\usepackage{arabtex}
\usepackage{utf8}
\usepackage{tikzsymbols}
\usepackage{amsmath}
\usepackage{graphicx}
\usepackage{multirow}
\usepackage{array,booktabs,makecell}
\usepackage{url}
\usepackage{enumitem}
\setenumerate[0]{label=\arabic*)}
\newcounter{examples}
\usepackage{wasysym}

\usepackage{multibib}
\newcites{languageresource}{Language Resources}
\usepackage{graphicx}
\usepackage{tabularx}
\usepackage{soul}

\newcommand\blfootnote[1]{%
  \begingroup
  \renewcommand\thefootnote{}\footnote{#1}%
  \addtocounter{footnote}{-1}%
  \endgroup
}

\usepackage{epstopdf}
\usepackage[utf8]{inputenc}

\usepackage{hyperref}
\usepackage{xstring}

\usepackage{color}
\makeatletter
\newcommand\footnoteref[1]{\protected@xdef\@thefnmark{\ref{#1}}\@footnotemark}
\makeatother

\title{Understanding and Detecting Dangerous Speech in Social Media}

\date{}
\name{Ali Alshehri$^{1\dagger}$, El Moatez Billah Nagoudi$^{2\dagger}$ , Muhammad Abdul-Mageed$^2$} 

\address{
$^1$ SUNY at Buffalo \\
$^2$ Natural Langauge Processing Lab, The University of British Columbia \\
        alimoham@buffalo.edu, \{moatez.nagoudi,muhammad.mageeed\}@ubc.ca}

\abstract{Social media communication has become a significant part of daily activity in modern societies. For this reason, ensuring safety in social media platforms is a necessity. Use of dangerous language such as physical threats in online environments is a somewhat rare, yet remains highly important. Although several works have been performed on the related issue of detecting offensive and hateful language, dangerous speech has not previously been treated in any significant way. Motivated by these observations, we report our efforts to build a labeled dataset for dangerous speech. We also exploit our dataset to develop highly effective models to detect dangerous content. Our best model performs at 59.60\% macro $F_1$, significantly outperforming a competitive baseline.}

\begin{document}
\maketitleabstract

\setcode{utf8}
\setarab
\novocalize

\section{Introduction}\label{sec:intro}
The proliferation of social media makes it necessary to ensure online safety. Unfortunately, offensive, hateful, aggressive, etc., language continues to be used online and put the well-being of millions of people at stake. In some cases, it has been reported that online incidents have caused not only mental and psychological trouble to some users but have indeed forced some to deactivate their accounts or, in extreme cases, even commit suicides~\blfootnote{ $^{\dagger}$ Both authors contributed equally.}\cite{hinduja2010bullying}. Previous work has focused on detecting various types of negative online behavior, but not necessarily dangerous speech. In this work, our goal is to bridge this gap by investigating dangerous content. More specifically, we focus on direct threats in Arabic Twitter. A threat can be defined as ``a statement of an intention to inflict pain, injury, damage, or other hostile action on someone in retribution for something done or not done."\footnote{\url{https://en.oxforddictionaries.com/definition/threat}} This definition highlights two main aspects: (1) the speaker's intention of committing an act, which (2) he/she believes to be unfavorable to the addressee~\cite{fraser98}. We especially direct our primary attention to threats of physical harm. We build a new dataset for training machine learning classifiers to detect dangerous speech. Clearly, resulting models can be beneficial in protecting online users and communities alike. \\

The fact that social media users can create fake accounts on online platforms makes it possible for such users to employ hostile and dangerous language without worrying about facing effective social nor legal consequences. This continues to put the responsibility on platforms such as Facebook and Twitter to maintain safe environments for their users. These networks have related guidelines and invest in fighting negative and dangerous content. Twitter, for example, prohibits any form of violence including threats of physical harm and promotion of terrorism.~\footnote{\url{https://help.twitter.com/en/rules-and-policies/twitter-rules}} However, due to the vast volume of communication on these platforms, it is not easy to detect harmful content manually. Our work aims at developing automated models to help alleviate this problem in the context of dangerous speech.\\

Our focus on Arabic is motivated by the wide use of social media in the Arab world~\cite{lenze2017social}. Relatively recent estimates indicate that there are over $11M$ monthly active users as of March $2017$, posting over $27M$ tweets each day~\cite{salem17}. An Arabic country such as Saudi Arabia has the highest Twitter penetration level worldwide, with 37\%~\cite{twitterStats}. The Arabic language also presents interesting challenges primarily due to the dialectical variations cutting across all its linguistic levels: phonetic, phonological, morphological, semantic and syntactic~\cite{farghaly2009arabic}. Our work caters for dialectal variations in that we collect data using multi-dialectal seeds (Section~\ref{sec:data}). Overall, we make the following contributions:

\begin{enumerate}
    \item We manually curate a multi-dialectal dictionary of \textit{physical harm threats} that can be used to collect data for training dangerous language models.
    \item We use our lexicon to collect a large dataset of threatening speech from Arabic Twitter, and manually annotate a subset of the data for dangerous speech. \textit{Our datasets are freely available online.}\footnoteref{note1} 
    \item We investigate and characterize threatening speech in Arabic Twitter.
    \item We train effective models for detecting dangerous speech in Arabic.
\end{enumerate}

The remainder of the paper is organized as follows: In Section~\ref{sec:rel}, we review related literature. Building dangerous lexica used to collect our datasets is discussed in Section~\ref{sec:data}. We describe our annotation in Section~\ref{sec:annot}. We present our models in Section~\ref{sec:models}, and conclude in Section~\ref{sec:con}. 

\section{Related work}\label{sec:rel}

Detection of offensive language in natural languages has recently attracted the interest of multiple researchers. However, the space of abusive language is vast and has its own nuances.~\newcite{Waseem17} classify abusive language along two dimensions: \textit{directness} (the level to which it is directed to a specific person or organization or not) and \textit{explicitness} (the degree to which it is explicit).~\newcite{jay2008pragmatics} categorize offensive language to three categories: \textit{Vulgar}, \textit{Pornographic}, and \textit{Hateful}. The Hateful category includes offensive language such as threats as well as language pertaining to class, race, or religion, among others. In the literature, these concepts are sometimes confused or even ignored altogether. In the following, we explore some of the relevant work on each of these themes.
\\

\textbf{Offensive Language.} The terms \textit{offensive language} and \textit{abusive language} are commonly used interchangeably. They are cover terms that usually include all types of undesirable language such as \textit{hateful}, \textit{racist}, \textit{obscene}, and \textit{dangerous} speech. We review some work looking at these types of language here, with no specific focus on any of its forms. GermEval 2018 is a shared task on the Identification of Offensive Language in German proposed by~\newcite{wiegand2018overview}. Their dataset consists of $8,500$ annotated tweets with two labels, ``offensive" and ``non-offensive". Another relevant shared task is the OffensEval~\cite{zampieri2019semeval}, which focuses on identifying and categorizing offensive language in social media. Very recently, an Arabic offensive language shared task is included in the 4th Workshop on Open-Source Arabic Corpora and Processing Tools (OSACT4).~\footnote{\url{http://edinburghnlp.inf.ed.ac.uk/workshops/OSACT4/}} \textit{}\\

\textbf{Hate Speech.} Hate speech is a type of language that is biased, hostile, and malicious targeting a person or a group of people because of some of their actual or perceived innate characteristics \cite{gitari2015lexicon}. This type of harmful language received the most attention in the literature.~\newcite{burnap2014hate} investigate the manifestation and diffusion of hate speech and antagonistic content in Twitter in relation to situations that could be classified as `trigger' events for hate crimes. Their dataset consists of $450K$ tweets collected during a two weeks window in the immediate aftermath of Drummer Lee Rigby’s murder in Woolwich, UK. In~\newcite{waseem2016you}, issues of annotation reliability are discussed. Authors examine  whether the expertise level of annotators (e.g  expert or amateur) and/or the type of information provided to the annotators, can improve the classification of hate speech. For this purpose, they extend the dataset of~\cite{waseem2016hateful} with a set of about $7K$ tweets annotated by two types of CrowdFlower users: expert and amateur. They find that hate speech detection models trained on expert annotations outperform those trained on amateur annotations. \textit{This suggests that hate speech can be implicit and thus harder to detect by humans and machines alike}. Another work by \cite{davidson2017automated} builds a hate speech lexicon based on a list of words and phrases provided by \textit{Hatebase.org}. Using Twitter API, they crawled a set of $85M$ tweets containing terms from the lexicon. Most recent works on detecting hate on Twitter are done as part of a SemEval2019 competition, HatEval~\cite{i2019multilingual}. This shared task addresses the problem of multilingual detection of hate speech against immigrants and women in Twitter. 
\\

\textbf{Obscene Language.} Obscene speech includes vulgar and pornographic speech. A few research papers have looked at this kind of speech in social media \cite{Singh2016,mubarak17,alshehri2018think}.~\newcite{mubarak17} present an automated method to create and expand a list of obscene words, for the purpose of detecting obscene language. \newcite{Abozinadah1} build a dataset of over $1M$ tweets comprising the most recent $50$ tweets of $255$ users who has participated in swearing hashtags as well as the most recent $50$ tweets of users in their network. As feature input to their classifiers, the authors extracted basic statistical measures from each tweet and reported $96\%$ accuracy of adult content detection.~\newcite{alshehri2018think} build a dataset of adult content in Arabic twitter and their distributors. The work identifies geographical distribution of targets of adult content and develops models for detecting spreaders of such content.\newcite{alshehri2018think} report 79\% accuracy on detecting adult content. \\

\textbf{Racism and Sexism.} \newcite{kwok2013locate} create a balanced dataset comprising $24,582$ of `racist' and `non-racist' tweets.~\newcite{waseem2016hateful} collect a set of $136K$ hate tweets based on a list of common terms and slurs pertaining ethnic minorities, gender, sexuality, and religion. Afterwards, a random set of $16K$ tweets are selected and manually annotated with three labels: `racist', `sexist', or ``neither".~\newcite{gamback2017using} introduce a deep-learning-based Twitter hate speech text classification model. Using data from~\newcite{waseem2016hateful} with about $6.5K$ tweets, the model classifies tweets into four categories: `sexist', `racist', `both sexist and racist', and `neither'.~\newcite{Clarke17}, using the same list, explore differences among racist and sexist tweets along three dimensions: \textit{interactiveness, antagonism}, and \textit{attitude} and find an overall significant difference between them. \textit{}\\

\textbf{Dangerous Language.}
Little work has been dedicated to detection and classification of dangerous language and threats. They are usually part of work on abusive and hate speech. This is to say that dangerous language has only been indirectly investigated within the NLP community. However, there is some research that is not necessarily computational in nature. For example, \newcite{gales11} investigates the correlation between interpersonal stance and the realization of threats by analyzing a corpus of $470$ authentic threats. Ultimately, the goal of Gale's work is to help predict violence before it occurs.~\newcite{hardaker16}, on the other hand, investigates the language surrounding threats of rape on Twitter. In their corpus, the authors find that women were the prime target of rape threats. In the rest of this paper, we explore the space and language of threats in Arabic Twitter. We now describe our lexicon and datasets. \\

\begin{table*}[t]
    \centering
    \begin{tabular}{c|l|l||c|l|l||c|l|l}
    \hline
    \textbf{Verb} & \textbf{Dialect} & \textbf{English} & \textbf{Verb} & 
    \textbf{Dialect} & \textbf{English} & \textbf{Verb} & \textbf{Dialect} &
    \textbf{English} \\ \hline
     \hline   
\<أباد>	& G,M,R	& exterminate	 & 	\<رض> & G,M	 & 	contuse	 & 	\<فجر> & all & blow up \\ 
\<أتل>	& E,L & kill & 	\<سطر> * & 	E,G	 & 	mark & \<فشق> & G,L &  split	 \\
\<أدى> * & 	E,G	 & 	give & 	\<سلخ> & all & skin & \<فقع>* & E,G,L,R	 & 	burst	 \\
\<أعدم> & all & execute	& \<سلق> & E,G,R & boil & \<فك>* & E,G,L & disentangle \\
\<أفنى>	& G,M,R	& exterminate & \<شج> & M & slash & \<قتل> & all & kill	 \\
\<أهلك>	& G,M,R & destroy	& \<شرب> ** & E,G,L,R	& drink & \<قرح> & E & sound \\
\<إغتال> & 	G,L,M,R	& assassinate & \<شق> & E,G,L,R	& rip off & \<قسم> & all & divide \\
\<إغتصب>	 & 	all	 & 	rape	 & 	\<شوه>	 & 	E,G,L,R	 & distort	& \<قصف> & G,R & smash \\
\<إقتلع> *	 & 	G,L,R	 & 	pluck	 & 	\<صرم>	 & 	G	 & 	cut off	 & 	\<قصم>	 & 	G,M	 & 	smash	 \\
\<بطش>	 & 	E,L,M	 & 	assault	 & 	\<صفق>	 & 	G,L	 & 	slap	 & 	\<قضى>	 & 	E,G,L,M	 & 	eliminate \\
\<جرح>	 & 	all	 & 	wound	 & 	\<صلخ>	 & 	G,L	 & 	skin	 & 	\<قطع>	 & 	all	 & 	cut	 \\
\<جزر>	 & 	G	 & 	cut off	 & 	\<ضرب>	 & 	all	 & 	hit	 & 	\<قلع>	 & 	E,G,L,R	 & 	pluck	 \\
\<جلد> & 	all	 & 	whip	 & 	\<طخ>	 & 	E,G	 & 	shoot	 & 	\<كسر>	 & 	all	 & 	break	 \\
\<حرق>	 & 	all	 & 	burn	 & 	\<طعن>	 & 	all	 & 	stab	 & 	\<لمخ>	 & 	G	 & 	hit	 \\
\<حطم>	 & 	E,L,M,R	& smash & \<طير> & E,G,L,R	& make fly & \<محا> ** & E,G,L,R	& erase \\
\<دك>	 & 	E,G,L & demolish & 	\<عذب>	 & 	E,G,M,R	 & 	torture	 & 	\<محق>	 & 	M	 & 	destroy	 \\
\<دهك>	 & 	G	 & 	run over & 	\<عزب>	 & 	E	 & 	torture	 & 	\<نحر>	 & 	E,G,M,R	 & 	slaughter	 \\
\<ذبح>	 & 	all & slaughter & \<عقر>	 & 	E,G	 & 	kill & 	\<نسف>	 & 	E,G,M,R	 & 	blast	 \\
\<رجم> & E,G,M,R	& stone & \<فتك> & E,G,L,M	& destroy	& \<هشم> & G,L,R & smash \\ \hline
    \end{tabular}
    \caption{Our list of dangerous verbs. \textit{NOTE.} All= all dialects, E= Egyptian, G= Gulf, L= Levantine, M= MSA. R= Maghrebi, * = metaphorical, ** = used idiomatically.}
    \label{verbs}
\end{table*}


\section{ Dangerous Lexica and Dataset}~\label{sec:data}
\vspace{-0.35cm}
\subsection{Dangerous Language}\label{sec:dang}

We define dangerous language as \textit{a statement of an intention to inflict physical pain, injury, or damage on someone in retribution for something done or not}. This definition excludes threats that do not reflect physical harm on the side of the receiver end of the threat. The definition also excludes \textit{tongue in cheek} whose real intention is to tease. An example of this later category is a threat made in the context of sports where it is common among fans to tease one another using metaphorical, string language (see Example \# 6 in Section~\ref{subsec:annot}).   

\subsection{Dangerous Lexica}\label{sec:lex}

We came up with a list of $57$ verbs in their basic form that can be used literally or metaphorically to indicate physical harm (see table~\ref{verbs}). This list is by no means exhaustive, although we did our best to expand it as much as possible. As such, the list covers the frequent verbs used in the threatening domain in Arabic.~\footnote{The concept of frequency here is based on native speaker knowledge of the language. The list was developed by the 3 authors, all of whom are native speakers of Arabic with multi-dialectal fluency.} These verbs are used in one or more of the following varieties: Egyptian, Gulf, Levantine, Maghrebi, and MSA (see table~\ref{dialect-verbs} for more details). Most of these verbs (n=$50$ out of $57$) literally indicate physical harm. Examples are \<طعن> (\textit{`to stap'}) and \<سلخ> (\textit{`to de-skin'}). The rest are used (sometimes metaphorically) to indicate threatening, such as \<اقتلع> (\textit{`to pluck'}) and \<سطر> (\textit{`to mark'}) usually with a body part such as  \<وجه> (\textit{`face'}) or  \<رأس> (\textit{`head'}). Finally, some of the verbs are used idiomatically, such as \<شرب من دم> (\textit{`to drink someone's blood'}) and \<محا من على وش الارض> (\textit{`to erase/eliminate from the face of the earth'}). Multiword expressions in our seed list can be found in Table \ref{multiwords}. \\ 

\begin{table}[ht]
    \centering
    \begin{tabular}{l|c}
    \hline
    \textbf{Dialect} & \textbf{\# of verbs}  \\ \hline   \hline
    MSA             & $30$ \\ 
    Gulf            & $50$ \\ 
    Egyptian        & $39$ \\ 
    Maghrebi        & $34$ \\ 
    Levantine       & $34$ \\ \hline 
    All (unique)    & $57$ \\ \hline
    \end{tabular}
    \caption{Distribution of threat verbs across Arabic dialects.}
    \label{dialect-verbs}
\end{table}

To be able to collect data, we used our manually curated list to construct threat phrases indicating physical harm such as \<اقتلك> (\textit{`I kill you'}) and \<يكسره> (\textit{`He breaks him/it'}). That is, each phrase consists of a physical harm verb, a singular or plural first or third person subject, and a plural or singular second or third person object. This gives us the following pattern:

\centerline{\emph{1st/3rd (SG / PL) + threat verb + 2nd/3rd (SG / PL)}} 
\bigskip

Some of the phrases only differ on the basis of spelling due to dialectical variations. For example, the body part \<وجه> (\textit{`face'}) can be spelled as \<وجيهكم> or \<وجوهكم> in the plural form depending on the dialect. Another example is the verb \<قتل> (\textit{`kill'}), which can also be spelled as \<اتل> in Egyptian and some other Arabic dialects. Manual search of some of the seed tokens in twitter suggests that patterns involving 3rd person subject are almost always not threats. The following are two illustrating examples of this non-threatening use:

\begin{enumerate}
\small   
  \setcounter{enumi}{\value{examples}}
  
      \item \<ميسي إذا لم يسجل فإنه يقتل فرحة بعض البشر> \\
    \textit{`If he doesn't score, Messi kills happiness in some people'}

    \item \<ما يكسر الخاطر ... سوى شخص غالي>  \\
    \textit{`Only a dear friend can break one's heart'}
    \setcounter{examples}{\value{enumi}}
\end{enumerate}

Thus, we decided to limit our list of phrases to `direct' dangerous threats, which are phrases involving a singular or plural first person subject and singular or plural second person object as follows:

\bigskip
\centerline{\emph{1st (SG/PL) + threat verb + 2nd (SG/PL)}}
\bigskip

Examples of these direct threats include \<نغتصبك> (\textit{`We rape you'}) and \<احرقكم> (\textit{`I burn you'}). Less dangerous threats such as \<اجرحكم> (\textit{``We hurt you (all)"}) and \<ادفك> (\textit{`I push you'}) are also not considered. Our motivation for not including these latter phrases even though they involve direct threats is that they indicate less danger and (more crucially) are more likely to be used metaphorically in Arabic. This resulted in a set of $286$ direct and dangerous phrases, which constitute our list of `dangerous' seeds. We make the list of $286$ direct threats phrases available to the research community.~\footnote{\label{note1}\url{https://github.com/UBC-NLP/ara_dangspeech}.}

\begin{center}
\begin{table*}[t]
\centering
\begin{tabular}{l|l||l|l}
\hline
\textbf{Seed} & \textbf{English} & \textbf{Seed} & \textbf{English} \\ \hline   \hline
\<	اشرب	من	دمك	>	&	I drink from your blood	&	\<	امحيك من على وش الارض			>	&	I erase you from the face of the earth	\\ 
\<	اشرب	دمك		>	&	I drink your blood	&	\<	امحيكم من على وش الارض			>	&	I erase you all from the face of the earth	\\ 
\<	اشرب	من	دمكم	>	&	I drink your blood all	&	\<	نشرب	من	دمك	>	&	We drink from your blood	\\ 
\<	اشوه	وجهك		>	&	I disfigure your face	&	\<	نشرب	دمك		>	&	We drink your blood	\\ 
\<	اطير	راسك		>	&	I cut your head	&	\<	نشرب	من	دمكم	>	&	We drink your blood all	\\ 
\<	اطير	روسكم		>	&	I cut your head all	&	\<	نطير	راسك		>	&	I cut your head	\\ 
\<	افجر	راسك		>	&	I blow up your head	&	\<	نطير	روسكم		>	&	I cut your head all	\\ 
\<	افقع	وجهك		>	&	I burst your face	&	\<	نفجر	راسك		>	&	We blow up your head	\\ 
\<	افقع	وجوهكم		>	&	I hit your face all	&	\<	نفقع	وجهك		>	&	We hit your face	\\ 
\<	افك	وجهك		>	&	I disentangle your face	&	\<	نقضي	عليك		>	&	We finish you	\\ 
\<	اقضي	عليك		>	&	I finish you	&	\<	نقضي	عليكم		>	&	We finish you all	\\ 
\<	اقضي	عليكم		>	&	I finish you all	&	\<	نمحيك من على وش الارض			>	&	I erase you from the face of the earth	\\
\<	اكسر	وجهك		>	&	I break your face	&	\<	نمحيكم من على وش الارض			>	&	We erase you all from the face of the earth	\\ \hline
\end{tabular}
\caption{Multiword expressions in our seed list.}
\label{multiwords}
\end{table*}
\end{center}

\begin{table}[ht]
    \centering 
        \begin{tabular}{l|c}
        \hline 
        \textbf{Dataset} & \# \textbf{of tweets} \\ \hline
        REST API    & $2.8M$   \\
        Timelines   & $107.5M$  \\ \hline \hline
        ALL          & $110.3M$  \\ \hline
        \end{tabular}
    \caption{Breakdown of our `dangerous' dataset.}
    \label{dang-table}
\end{table}

\vspace{-1.25cm}
\subsection{Dataset}\label{sec:dataset}

We use the constructed `dangerous' seed list to search Twitter using the REST API for two weeks resulting in a dataset of $2.8M$ tweets involving `direct' threats as shown in Table~\ref{dang-table}. We then extract \textit{user ids} from all users who contributed the REST API data ($n=399K$ users) and crawled their timelines ($n=705M$ tweets). We then acquire $107.5M$ tweets from the timelines, each of which carry one or more items from our `dangerous' seed list. 
Combining these two datasets (the REST API dataset and dataset based on the timelines) results in a dataset consisting of $110.3M$ tweets as shown in Table \ref{dang-table}. In this work, we focus on exploiting the REST API dataset exclusively, leaving the rest of the data to future research.

\section{Data Annotation}~\label{sec:annot}
\vspace{-0.4cm}
\subsection{Annotation}\label{subsec:annot}

We first randomly sample $1K$ tweets from our REST API dataset.\footnoteref{note1} Two of the authors annotated each tweet for being a threat (`dangerous') or not (`safe'). This  sample annotation resulted in a Kappa ($\kappa$) score of $0.57$, which is fair according to Landis and Koch’s scale~\cite{kochs}. The two annotators then held several discussion sessions to improve their mutual understanding of the problem and define some instructions as to how to label the data. We also added another random sample of $4K$ tweets (for a total size of $5K$) to the annotation pool. After extensive revisions of the disagreement cases by the two annotators, the $\kappa$ score for the whole dataset ($5K$) was found to be at $0.90$. The annotated dataset has a total of $1,375$ tweets in the `dangerous' class and $3,636$ in the `non-dangerous' class. Our overall agreed-upon instructions for annotations include the following:

\begin{table}[ht]
    \centering
        \begin{tabular}{l|c|c|c}
        \hline 
        \textbf{} & \textbf{Safe} & \textbf{Dangerous} & \textbf{Total} \\ \hline \hline
        \textbf{Safe}           & $3,570$   & $ 52 $   & $3,622$  \\ 
        \textbf{Dangerous}      & $ 70 $   & $1319$   & $1,389$ \\
        \textbf{Total}          & $3,640$   & $1,371$   & $5,011$ \\ \hline
        \end{tabular}
    \caption{Annotator Agreement of $5011$-tweet sample.}
    \label{agree}
\end{table}

\begin{itemize}
    \item Textual threats combined with pleasant emojis such as \includegraphics[height=1.2em]{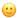} and \includegraphics[height=1.2em]{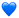} are not dangerous, as opposed to threat combined with less pleasant emojis such as \includegraphics[height=1.2em]{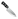} \includegraphics[height=1.2em]{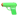}. Thus, tweet $3$ below should be coded as `safe' while tweet $4$ should be tagged as `dangerous'. 
    
    \begin{enumerate}
 \small    
    \setcounter{enumi}{\value{examples}}
    \item @user \<المنطق يقول أنا باقتلك> \includegraphics[height=1.2em]{smile.png}\\
    \textit{`It goes with logic that I kill you \includegraphics[height=1.2em]{smile.png}'}\\
    
    \item @user @user @user \<قدامي بس لا اطعنك> \includegraphics[height=1.2em]{nife.png}\\
    \textit{`Move forward [in front of me] or else I stab you \includegraphics[height=1.2em]{nife.png}'} \\
    \setcounter{examples}{\value{enumi}}
    \end{enumerate}

    \item Mitigated threats with question marks or epistemic modals are dangerous unless they are combined with positive language or emojis such as Example $5$ below. Note that the word \textit{Touha} in Example 5 is an informal, friendly form for Arabic names such as \textit{FatHi} or \textit{MamdouH}.
    
    \begin{enumerate}
    \small
    \setcounter{enumi}{\value{examples}}
    \item @user \<انا بفكر اقتلك يا توحه >  \includegraphics[height=1.2em]{smile.png}\\
    \textit{`I am thinking of killing you, Touha \includegraphics[height=1.2em]{smile.png}'}\\
    \setcounter{examples}{\value{enumi}}
    \end{enumerate}

    \item Threats related to sports are not dangerous. That is because it is common to use verbs like \<نحر> (\textit{``slaughter"}) and \<اغتصب> (\textit{``rape"}) among fans of rival teams to describe wins and losses, as in the following example.
    
    \begin{enumerate}
    \small
      
    \setcounter{enumi}{\value{examples}}
        \item @user \< حلاته نغتصبكم على ارضكم وبين جمهوركم >\\
    \textit{`It's actually better that we `rape' you in your stadium, among your fans'}\\
    \setcounter{examples}{\value{enumi}}
    \end{enumerate}

    \item Ambiguous threats such as threats consisting of one word (as in Example 7 below) should be coded as `dangerous':
    
    \begin{enumerate}
    
      \small
    \setcounter{enumi}{\value{examples}}
    \item \<اقتلكوا> \\
    \textit{`I kill you'}\\
    \setcounter{examples}{\value{enumi}}
    \end{enumerate}
    
\end{itemize}

Below, we show examples of tweets that were annotated as `dangerous':

\begin{enumerate}
    \small
    \setcounter{enumi}{\value{examples}}
    \item @user @user \\ \<ودي احرقك و ارميك للكلاب> \\
    \textit{`I wish to burn you and throw you to dogs'}\\

    \item @user \<ماتكلمنيش ب الطريقة دي لحسن اقوم اضربك > \\
    \includegraphics[height=1.3em]{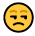} \< قال اربع نسوة قال> \ \\
    \textit{`Don't talk to me in this way, or else I hit you! Talking of (marrying) four women!'}\\


\item @user @user \<سوف تبدا الحرب ورب العرش العظيم  سوف >\\
    \<نحرقكم حرق انتم يامخنثييين يا خنزير العرب يا خاونه>\\
    \textit{`The war will begin. By God, we will burn you down, you fags, you pigs, you traitors'}

\item @user \<الحمار دايم حمار مااستفدتوا من الدرس هاذا لازم  > \\
    \<نضربكم على قفاكم زي الجهال انتوا بشر ولاحيونات> \\    
    \textit{`A donkey will always be a donkey. You didn't learn the lesson. We have to hit you on the back of you heads like kids. Are you humans or animals?'}\\

    \item @user \<عطيني كروكي بيتكم واجي اشرحك مو بس اقتلك> \\
    \textit{`Give me your address so I can come to you, and not only kill you but also dissect you'}
    \setcounter{examples}{\value{enumi}}
    
\end{enumerate}

\begin{table}[ht]
    \centering
        \begin{tabular}{l|c}
        \hline 
        \textbf{Measure} & \textbf{Value}  \\ \hline \hline
        Avg. \# timeline tweets & $2,313$    \\
        Avg. \# dangerous tweets / user & $3.97$   \\
        St. dev.   & $3.64$   \\ 
        25th percentile       & $1$       \\ 
        50th percentile       & $4$       \\ 
        75th percentile       & $6$       \\ 
        Minimum    & $1$       \\ 
        Maximum    & $23$      \\ \hline

        \end{tabular}
    \caption{Descriptive statistics of the timeline data of $1,370$ users who contributed tweets classified as `dangerous' in our annotated dataset.}
    \label{user-dang}
\end{table}

\begin{center}
\begin{table}[t]
\small
   \centering
\begin{tabular}{l|l||l}

\hline

\textbf{Seed}&  \textbf{English} & \textbf{Emoji}
 \\ \hline \hline
 
\<	اذبحك	>	&	I slaughter you	&	\includegraphics[height=1.4em]{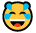}   \\ 
\<	اقتلك	>	&	I kill you	&	\includegraphics[height=1.4em]{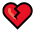}   \\ 
\<	اغتصبك	>	&	I rape you	&	\includegraphics[height=1.4em]{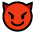}   \\ 
\<	اضربك	>	&	I hit you	&	\includegraphics[height=1.4em]{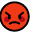}   \\ 
\<	اعذبك	>	&	I torture you	&	\includegraphics[height=1.5em]{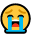}   \\ 
\<	اديك	>	&	I hit/give you	&	\includegraphics[height=1.4em]{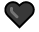}   \\ 
\<	اجلدك	>	&	I lash you	&	\includegraphics[height=1.4em]{image/em7.PNG} \\ 
\<	اطعنك	>	&	I stab you	&	\includegraphics[height=1.4em]{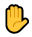}   \\ 
\<	اجرحك	>	&	I hurt you	&	\includegraphics[height=1.4em]{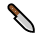}   \\ 
\<	احرقك	>	&	I burn you	&	\includegraphics[height=1.4em]{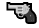}   \\ \hline

\end{tabular}
\caption{Top $10$ most frequent `dangerous' seeds and emojis in our REST API dataset.}
\label{top-t0}
\end{table}
\end{center}

\begin{table*}[ht]
\centering
\begin{tabular}{lccccc}
 \toprule
  
\textbf{Models}  & \textbf{Datasets}  &  \bf Precision &\bf Recall &\bf Acc  &  \bf F$_{1}$ \\ \midrule \midrule
\bf Baseline  & -- &   $50.00$ & $29.33$    &  $ 58.66$  & $36.97$  \\
\toprule
BERT  & Dangerous   &    $58.42$ & \bf 60.10    &  $ 74.27$ & \ $58.98$\\
BERT  & Dangerous + Offensive  &   $53.80$ &  $53.44$     &  $66.11$ & \ $53.52$ \\ 

\midrule
\midrule
BERT-Emotion  & Dangerous  &   \bf60.06 &  $59.24$    &  \bf 77.97   &   \bf 59.60 \\ 

BERT-Emotion  &  Dangerous + Offensive   & $54.50$ &    $53.99$  &  $ 66.84$ &  $54.11$ \\

\bottomrule

\end{tabular}
\caption{Results from our models on TEST.}
\label{tab:res}
\end{table*}

\subsection{Data Analysis}\label{sec:analysis}

The fact that `dangerous' tweets are not frequent in the dataset suggests that \textit{this phenomenon of dangerous speech is relatively rare in the Twitter domain}. To further investigate the commonality of such a phenomenon, we extract the timelines of the authors of tweets in the dangerous class in the annotated dataset. Table~\ref{user-dang} shows some descriptive statistics of the occurrence of dangerous seeds in their timelines. We can see from Table~\ref{user-dang} that timelines contain on average $2,313$ tweets for each user, and there are on average $3.97$ tweets in each timeline containing a dangerous seed token. This represents $\sim 0.17\%$ of the tweets for each user. The average number of dangerous tweets is higher ($n= 6$) for users in the $75th$ percentile as opposed to $n= 1$ in the $25th$ percentile.

To further understand dangerous language, we also analyze all the $5,011$ tweets from our annotated dataset. We identify a number of patterns in the data, cutting across both the `dangerous' and `safe' classes. We explain each of these patterns next. \\

\noindent{\bf Conditional threats:} One common threatening pattern involves conditional statements where the consequent involves a physical threat by the speaker toward the addressee, and the antecedent is a conditional phrase involving deterrence of an action that can possibly be carried out by the addressee or someone else. The following are two examples:

\begin{enumerate} \label{ex11}
    \small
    \setcounter{enumi}{\value{examples}}
    \item @user \< اذبحك اذا تسوين شى> \\
    \textit{`I slaughter you if you (F) do anything'}

    \item @user  \<اذا انتقل سوف اطعنك طعناً مبرحاآ امام الملآ> \\
    \textit{`If he transfers, I will stab you hardly in front of the crowds'}
    \setcounter{examples}{\value{enumi}}
    
\end{enumerate}
  
It is clear from Examples 13 and 14 that the threats are directed to a twitter user mentioned in the tweet. So these tweets are potentially part of ongoing conversations between the person who posted the tweet and the user mentioned in the body of the tweet. As Table~\ref{counts} shows, $\sim 71.2$\% of tweets in our annotated dataset (across the `dangerous' and `safe' classes) contain mentions of other Twitter users. This percentage is higher within the dangerous class (\%= $78$). \\

\noindent{\bf Threats accompanied with commands:} Another common pattern involves a command accompanying the threat as in Example 15 below. These kinds of threats are more common in the dangerous than the safe class.

\begin{enumerate}

    \setcounter{enumi}{\value{examples}}
    \item @user \< اقول انقلعي لا افقع وجهك>\\
    \textit{`I say get out before I hit your face'}
    \setcounter{examples}{\value{enumi}}
\end{enumerate}

\noindent{\bf Threats accompanied with questions:} Another less common pattern is threats in the form of questions. This kind of threats occurs in about $5$\% of our dangerous data as compared to $2.8$\% in the safe class. Unlike the examples above, the reason behind most of the `question' threats is not particularly clear as they tend to be short, sometimes of one word. Interpretation of these threats requires more context, beyond the level of the tweet itself. Examples 16-18 illustrate this category. 

\begin{enumerate}
\small
    \setcounter{enumi}{\value{examples}}
    \item \<ممكن اقتلك من الم الحبر؟> \\
    \textit{`can i kill you by the pen'}

    \item @user \<ينفع اغتصبك؟> \\
    \textit{`Does it work if I rape you?'}
    
    \item @user \<اذبحك؟> \\
    \textit{`I slaughter you?'}
    \setcounter{examples}{\value{enumi}}
\end{enumerate}
 
\noindent{\bf Threat accompanied by modality:} Some threats carry \textit{deontic modality} where modals such as `would', `probably', `may' are employed. \textit{Epistemic modality} are also found in some data points. Similar to the question types above, these tweets (Examples 19-21 below) are less threatening than Examples 13-18 above. 

\begin{enumerate}
\small
    \setcounter{enumi}{\value{examples}}

    \item @user \<جعلني اغتصبك> \\
    \textit{`May I rape you?!'}
    \setcounter{examples}{\value{enumi}} (\textbf{\textit{deontic modalit}y})
    
    \item @user \<ودي اذبحك> \\
    \textit{`I would like to kill you'} (\textbf{\textit{deontic modalit}y})
    
    \item @user \<شكلي رح اقتلك مع صحبتك> \\
    \textit{`I am probably going to kill you with your friends'} (\textbf{\textit{epistemic modality}})

\end{enumerate}

\begin{table}[t]
    \centering
        \begin{tabular}{p{1.7cm}|p{1.3cm}|p{1.6cm}|p{1.7cm}}
        \hline
        \textbf{Phenomena} & \textbf{Freq.} & \textbf{Percentage (non-dangerous)} & \textbf{Percentage (dangerous class)} \\ \hline \hline
        Mentions    & $3673$ & $72.8$\% & $78$\% \\
        Questions   & $100$ & $2.8$\%  & $5$\%  \\
        Emoji       & $2,010$ & $45.5$\%  & $36$\% \\
        Conditional &   $742$ & $15.8$\% & $11.4$\%   \\ 
        Body parts  &   $378$ & $6.6$\% &   $11.3$\% \\
        Hahaha      & $355$ &  $9.9$\%  & $1.1$\% \\\hline
        \end{tabular}
    \caption{The frequency of some textual phenomena in our Annotated data.}
    \label{counts}
\end{table}

\noindent{\bf Metaphorical threats:} Many of the tweets involve metaphorical use of the phrases in our annotated data. The target domain of the majority of these metaphorical uses is either sports or relationships. Words such as `kill', `rape', and `slaughter' are used to indicate `wining' in sport or `burn' to mean `pain' or `longing' in romantic relationships. Examples 23-24 illustrate these cases: 

\begin{enumerate}
\small
    \setcounter{enumi}{\value{examples}}
    \item \<احب قول لاخواني المانشستراويه بكره راح>\\
    \<نغتصبكم فلا داعي للذعر والضجر وانمااستمتعوا> \\
    \textit{`I would like to tell my Manchester (football club) fans that we will rape them tomorrow'}

    \item  \<س احرقك عشقا واطفئك غراما> \\
    \textit{`I will burn you with love and put off (the fire on you) with affection'}
    \setcounter{examples}{\value{enumi}}
\end{enumerate}

\noindent{\bf Emojis:} Another interesting phenomenon (see Table \ref{counts}) is the frequent use of emojis, which are found in about $40$\% of the annotated dataset. This is not surprising as it helps participants mitigate (and hence better disambiguate the nature of) their threats. Table~\ref{top-t0} shows the top most frequent emojis used in our REST API data. It is evident that most of the used emojis do not indicate friendliness, but rather have a threatening nature. This is also true of using expressive interjections such as \textit{hahaha}, which is more common in the non-dangerous than the dangerous class. Additionally, as mentioned above, some expressions involve use of `body parts' such as \textit{eyes, head, face, nose}, etc.. These are found to occur significantly higher in the `dangerous' class. \\

\noindent{\bf Conversational context:} Finally, Table \ref{top-t0} also shows the top 10 most frequent seeds in our REST API dataset. All of these seeds involve a first singular person subject and a singular second person object, which indicate that many of these tweets containing dangerous seeds are part of one-to-one conversations on Twitter. 
\vspace{0.4cm}

\section{Deep Learning Models}\label{sec:models} 
\vspace{0.3cm}

\textbf{Dangerous speech data.} We use our $5,011$ annotated tweet dataset for training deep learning models on dangerous speech. The dataset comprises $3,570$ `safe' tweets and $1,389$ `dangerous' tweets.  We first remove all the seeds in our lexicon since these were used in collecting the data. We then keep only tweets with at least two words, obtaining $4,445$ tweets with $3,225$ `safe' labels and $1,220$ `dangerous' tweet (see Table~\ref{split-data}). We split this dataset into $80\%$ training, $10\%$ development, and $10\%$ test. \\ 

\textbf{Offensive speech data.} In one of our settings, we also use the offensive dataset released via the Offensive Shared Task $2020$.\footnote{\url{http://edinburghnlp.inf.ed.ac.uk/workshops/OSACT4/}} This offensive content dataset consists of $8000$ tweets ($1,590$ `offensive' and $6,410$ `non-offensive'). We use the offensive class data to augment our train split. Hence, we evaluate only on our test split where tweets are restricted to our dangerous gold tweets in the annotated dataset. We run this experiment as a way to test the utility of exploiting offensive tweets for enhancing dangerous language representation based on the assumption that dangerous speech is a subset of offensive language. However, as we see in Table~\ref{tab:res}, this measure did not result in any improvements on top of our dangerous models. In fact, it leads to model deterioration.\\

\begin{table}[ht]
    \centering
        \begin{tabular}{l|c|c|c}
        \hline 
         \textbf{} &  \textbf{Train} & \textbf{Dev} & \textbf{Test} \\ \hline \hline
        \textbf{\#Safe}        & $2,727$   & $ 244 $   & $254$  \\ 
        \textbf{\#Dangerous}      & $ 852 $   & $189$   & $179$ \\ \hline
        \textbf{Total}         & $3,579$   & $433$   & $433$ \\ \hline  
        \end{tabular}   
    \caption{Distribution of dangerous and safe  classes in our annotated dataset after normalization by removing seeds and one-word tweets.}
    \label{split-data}
\end{table}

\textbf{Models.} For the purpose of training deep learning models for detecting dangerous speech, we exploit the Bidirectional Encoder Representations from Transformers (BERT) \cite{devlin2018bert} model. For all our models, we use the BERT-Base Multilingual Cased (Multi-Cased) model.\footnote{\url{https://github.com/google-research/bert/blob/master/multilingual.md}.} It is trained on Wikipedia for $104$ languages (including Arabic) with  $12$ layers, $12$ attention heads, $768$ hidden units each and $110$M parameters. Additionally, we further fine-tune an off-the-shelf trained BERT Emotion (BERT-EMO) from AraNet~\cite{abdul2019aranet} on our dangerous speech task. BERT-EMO is trained with Google's BERT-Base Multilingual Cased model on $8$ emotion classes exploiting Arabic 
Twitter data. We train all BERT models for $20$ epochs with a batch size of $32$, maximum sequence size of $50$ tokens and learning rate up to $2e^{-5}$. We identify best results on the development set, and report final results on the blind test set. As our baseline, we use the majority class in our training split. Note that since our dataset is not balanced, the majority class baseline is competitive (63.97\% macro $F_{1}$ score). Also, importantly, due to the imbalance in class distribution, the macro $F_{1}$ score (the harmonic mean of precision and recall) is our metric of choice as it is more balanced than accuracy.\\

\textbf{Results \& Discussion.}  
As Table~\ref{tab:res} shows, the results demonstrate that all the models outperform the baseline and succeed in detecting the dangerous speech with $F_{1}$ scores between $53.42\%$ and $59.60\%$. We also observe that training on the offensive dataset did not improve the results. On the contrary, augmenting training data with the offensive task tweets cause deterioration to $53.52$\% $F_{1}$ for BERT and $54.11$\% $F_{1}$ for BERT-Emotion. \\

The best model for detecting dangerous tweets is BERT-Emotion when fine-tuned on our gold dangerous dataset. It obtains an accuracy level of $77.97$\% and $F_{1}$ score of $59.60$\%. We note that \textit{both} accuracy and $F_{1}$ are \textit{significantly higher then the the baseline}. As mentioned earlier, since our dataset is highly imbalanced, $F_{1}$, rather than accuracy, should be used as the metric of choice for evaluation. As such, our models are significantly better than our baseline.

\section{Conclusion}\label{sec:con}
\vspace{0.5cm}
We have described our efforts to collect and manually label a dangerous speech dataset from a range of Arabic varieties. Our work shows that dangerous speech is rare online, thus making it difficult to find data for training machine learning classifiers. However, we were able to collect and annotate a sizeable dataset. To accelerate research, we will make our data available upon request. Another contribution we made is developing a number of models exploiting our data. Our best models are effective, and can be deployed for detecting the rare, yet highly serious, phenomenon of dangerous speech. For future work, we plan to further explore contexts of use of dangerous language in social media. We also plan to explore other deep learning methods on the task. 

%

\section*{Acknowledgements}
We acknowledge the support of the Natural Sciences and Engineering Research Council of Canada (NSERC), the Social Sciences Research Council of Canada (SSHRC), and Compute Canada (\url{www.computecanada.ca}).
\section{Bibliographical References}

\bibliography{lrec2020.bib}
\bibliographystyle{lrec.bst}

\end{document}